\documentclass[wcp]{jmlr}

\usepackage{longtable}% for long tables

\usepackage{graphicx}
\usepackage{xcolor}
\usepackage{amsmath}
\usepackage{multirow}
\usepackage[justification=centering]{caption}
\usepackage{url}
\usepackage{booktabs}
\usepackage{adjustbox}

\usepackage{booktabs}
\makeatletter
\def\set@curr@file#1{\def\@curr@file{#1}} %temp workaround for 2019 latex release
\makeatother
\usepackage[load-configurations=version-1]{siunitx} % newer version

\jmlrvolume{xx}
\jmlryear{2020}
\jmlrworkshop{ACML 2020}

\title[Proxy Network]{Proxy Network for Few Shot Learning}

\author{\Name{Bin Xiao} \Email{buptxiaofeng@gmail.com}\\
  \addr Department of Computer Science, National Chiao Tung University, 1001 University Road, Hsinchu 30010, Taiwan, ROC
  \AND
  \Name{Chien-Liang Liu} \Email{clliu@mail.nctu.edu.tw}\\
  \addr Department of Industrial Engineering and Management, National Chiao Tung University, 1001 University Road, Hsinchu 30010, Taiwan, ROC
  \AND
  \Name{Wen-Hoar Hsaio} \Email{bass28.cs96g@g2.nctu.edu.tw}\\
  \addr Information Management Center, National Chung-Shan Institute of Science and Technology, Taoyuan 32546, Taiwan, ROC
 }

\begin{document}

\maketitle

\begin{abstract}
The use of a few examples for each class to train a predictive model that can be generalized to novel classes is a crucial and valuable research direction in artificial intelligence. This work addresses this problem by proposing a few-shot learning (FSL) algorithm called proxy network under the architecture of meta-learning. Metric-learning based approaches assume that the data points within the same class should be close, whereas the data points in the different classes should be separated as far as possible in the embedding space. We conclude that the success of metric-learning based approaches lies in the data embedding, the representative of each class, and the distance metric. In this work, we propose a simple but effective end-to-end model that directly learns proxies for class representative and distance metric from data simultaneously. We conduct experiments on CUB and mini-ImageNet datasets in 1-shot-5-way and 5-shot-5-way scenarios, and the experimental results demonstrate the superiority of our proposed method over state-of-the-art methods. Besides, we provide a detailed analysis of our proposed method. 
\end{abstract}

\section{Introduction}
The last decade has witnessed great success of deep learning~\citep{krizhevsky2012imagenet,lecun2015deep}, because it has achieved promising results in many application domains. In image domain, convolutional neural networks~\citep{lecun1998gradient,krizhevsky2012imagenet,lecun2015deep,he2016deep} (CNNs) can benefit from their shared-weights architecture and translation invariance characteristics to learn discriminative and hierarchical features from data without human intervention. However, thousands of labeled examples for each class are required to saturate performance in classification problems, but this differs from human learning. Humans can recognize new object classes even though only few instances are available at hand. Thus, the use of a few examples for each class to train a predictive model that can be generalized to novel classes is a crucial and valuable research direction.

Many recent works proposed to use few-shot learning (FSL) and meta learning~\citep{fei2006one,fink2005object,triantafillou2017few,snell2017prototypical,finn2017model} to deal with aforementioned problem. A traditional machine learning algorithm aims to learn and optimize a single task, whereas a meta-learning model is trained over a variety of learning tasks and optimized for the best performance on a distribution of tasks, including potentially unseen tasks~\citep{weng2018metalearning}. Notably, the problem pertaining to how to quickly learn a model for generalizing it to a new task is the core objective of meta learning, explaining why meta learning is also known as learning to learn. 

%Notably, meta learning aims to learn how to learn, so that the model can learn to adapt to new tasks rapidly with only few instances are available for the training of new tasks.  

Many previous FSL works~\citep{vinyals2016matching,snell2017prototypical,sung2018learning} have employed the meta-learning architecture to train and evaluate the model by using various tasks. An immediate model can be obtained through learning from a variety of tasks, and the learned model can quickly adapt to a new task by using a few data examples~\citep{finn2017model}. Overcoming this problem is also crucial for the development of artificial intelligence to quickly learn new things with prior knowledge as humans do. Similarly, the FSL models can use the prior knowledge that has learned from various tasks to constrain the search space of the new task~\citep{wang2019few}. 

Metric-based approaches are a class of methods that can be used to attack FSL problems, and the goal is to learn an embedding space that makes the data points in the same class to be close, whereas the data points in different classes to be separated as far as possible. We analyze the previous metric-based approaches on FSL problems and conclude that the success of metric-based approaches lies in three components: data embedding, the representative of each class, and the distance metric. We propose to use data-driven approaches to develop a novel model called proxy network (ProxyNet) under meta-learning settings. In our proposed ProxyNet, the data embeddings, and the proxies for class representatives as well as distance metric are all learned from data, and these differ from the previous metric-based approaches on FSL. The proposed model is simple to implement, but the experimental results demonstrate the superiority of our proposed method over state-of-the-art methods on 1-shot-5-way and 5-shot-5-way problems.

%learn the three elements from data, so two key differences exist between the proposed method and the previous metric-based methods on FSL. First, we proposed to learn the class representative for each class and used it as the proxy of the class. Second, we proposed to directly learn the distance metric from the data and used it as the proxy of the distance metric. 

The contributions of this work are listed as follows. First, we analyze the previous metric-based approaches on FSL problems and conclude the key elements that lead to the success of metric-based approaches. Then, we propose a novel FSL algorithm called ProxyNet based on our analysis and the meta-learning settings. Second, we conduct experiments on two well-known benchmark datasets, and compare our proposed method with recent state-of-the-art methods. The experimental results indicate that our proposed method significantly outperforms these methods. Finally, we provide detailed analysis and investigation about our proposed method.

\section{Related Work}
Few-shot learning  has attracted much attention in recent years as the goal is to recognize novel classes by using very few training examples~\citep{fei2006one,vinyals2016matching,snell2017prototypical,finn2017model,sung2018learning}. Notably, a typical CNN architecture comprises feature extractor and a classifier, and several previous works proposed to modify the classifier to deal with novel classes with a few examples. For example, \citet{qi2018low} proposed an imprinting method to directly set the weights of final layer for novel classes from the embeddings of training exemplars. \citet{gidaris2018dynamic} proposed to extend CNN with an attention-based FSL weight generator and used cosine similarity between feature representations and classification weight vectors to determine prediction outcomes. \citet{wang2019few} conducted a detailed survey about recent FSL works and categorized these methods into several approaches, such as metric-based, optimization-based, and augmentation-based methods.

Metric-based algorithms address the few-shot classification problem by learning to compare. \citet{koch2015siamese} combined a siamese architecture~\citep{chopra2005learning} and a convolutional architecture to develop deep siamese networks for learning data representations and modeling few-shot learning as a verification task. Different from traditional deep learning models, the siamese network receives two inputs every time, and the two inputs are fed into the same network architecture with the same weights to transform the inputs into an embedding space. Once the transformation is completed, the objective is to make the data points that are in the same class to be close and the data points in different classes to be far away. \citet{hoffer2015deep} extended the idea of siamese networks to devise triplet networks that accept three data points as the inputs at a time, including anchor data (reference point), positive data, and negative data. The aim of triplet loss is to bring the reference point near the positive point and away from the negative point.

%Although siamese networks and triplet networks can classify the data points that are in the novel classes, they do not use meta-learning framework to train the models and only consider pairwise comparison during model training. 

Matching networks~\citep{vinyals2016matching} were the first to employ the meta-learning architecture that trains and tests using the setting of $K$-shot-$N$-way tasks, in which $N$ is the number of classes in each task, and $K$ represents the labeled examples per class. Matching networks mapped a small number of labeled examples in the support set and unlabeled examples in the query set to embedding spaces with different embedding architectures. Cosine similarity between the embeddings of the data points in the support and query sets were used to determine the classification outcome. \citet{snell2017prototypical} proposed prototypical networks by using the mean embedding or prototype as the representation for each class. The Euclidean distance between query embedding and the mean embedding for each class becomes the basis of the classification. Different from matching networks and prototypical networks that use pre-defined distance metrics to measure the similarity between the embeddings, the relation network model proposed by \citet{sung2018learning} comprises two parts, embedding and relation modules. The purpose of embedding module was the same as those of aforementioned metric-based FSL algorithms, while the relation module used a learnable distance metric to calculate the similarity of embeddings by obtaining relation scores. Notably, the prediction of relation scores is considered a regression problem even though the targets of ground-truth are the values of 0 or 1.

Optimization-based methods constrain the learning algorithms by providing a good initialization of the model parameters, or learning the update rules. Model agnostic meta-learning (MAML)~\citep{finn2017model} is a meta-learning framework that learns a network initialization that can quickly adapt to new tasks. \citet{nichol2018first} introduced a variant of first order MAML called Reptile and explored the potential of meta-learning algorithms based on first-order gradient information. Variants of gradient-based optimization algorithms are standard optimization algorithms to train deep neural networks. \citet{ravi2016optimization} discovered that the update rule for the variants of gradient descent resembles the update rule for the cell state in a long short-term memory (LSTM)~\citep{hochreiter1997long}, inspiring them to propose an LSTM-based meta-learner optimizer that was trained to optimize a NN classifier. 

Augmentation-based methods cope with FSL by hallucinating new examples to augment datasets for the novel classes. Intuitively, this approach relies on the quality of the generated examples. \citet{hariharan2017low} considered to use analogy relationship to generate new examples. Given quadruplets of features taken from training tasks, the aim was to train a generator that can output a hallucinated feature vector by transferring modes of variation from the base classes. Subsequently, \citet{wang2018low} proposed a method that jointly trained a meta-learner and a hallucinator, which mapped real seed examples and noise vectors to hallucinated examples.

%by using a convolutional NN (CNN) architecture. Notably, to perform classification, the relation network employs a learnable distance metric that is based on the CNN in the relation module rather than using a predefined distance metric, such as Euclidean or cosine distance metric, used in related prior few-shot studies. Moreover, the objective function of the relation network is a mean squared error (MSE) loss function because the relation network focuses more on predicting relation scores, which resembles regression to a greater extent than classification. 

%for few-shot learning based on the idea that there exists an embedding in which points cluster around a single prototype representation provided for each class. Prototypical networks conduct few-shot classification by computing distances between the prototype representations of each class. The network recognizes queries that it has never seen during training, and the support set comprises few samples for each category. 

\section{Proposed Method}
This section briefly introduces FSL problems in the meta-learning framework. Next, we analyze the components involved in the previous metric-based methods on FSL and present an overall architecture as well as these key components for metric-based methods. Based on this architecture, we further present our proposed model, proxy networks (ProxyNet).

\subsection{Problem Definition}
In this study, we follow meta-learning settings to design the model and conduct experiments. First, we separate the process into meta-training and meta-testing stages. Notably, the datasets in meta-training and meta-testing have disjoint classes as the purpose is to evaluate whether the model can make accurate predictions on novel classes. Second, the training and evaluation are conducted with a collection of $K$-shot-$N$-way tasks in episodes. Conventionally, $K$ is less than 20 in the setting of FSL, and the goal is to assess whether the model can quickly adapt to new tasks with few labeled examples. 

%% temporily comment
In meta-training stage, each task comprises a support set, $S = \left\{(\mathbf{x}_i,y_i)\right\}_{i=1}^{N \cdot K}$, where $\mathbf{x}_i$ is the input example and $y_i \in \{1, \ldots, N\}$ is the corresponding label, and a query set, $Q = \left\{(\mathbf{x}_j,y_j)\right\}_{j=1}^{N \cdot T}$, where $T$ is the number of  query instances per class in each meta-training episode, $\mathbf{x}_j$ is the query example and $y_j \in \{1, \ldots, N\}$ is the corresponding label. Additionally, $S \cap Q = \O$, but $S$ and $Q$ share the same label space. The algorithm has to determine which of the support set class $c$ the query samples $q$ belongs to based on the class assignment probability $P_{\phi}(y=c|q)$, in which $\phi$ is model parameter. The settings of meta-testing resemble those of the meta-training stage except that the query set dose not comprise label $y$.

\begin{figure}[htp]
\begin{center}
  \includegraphics[width=\textwidth]{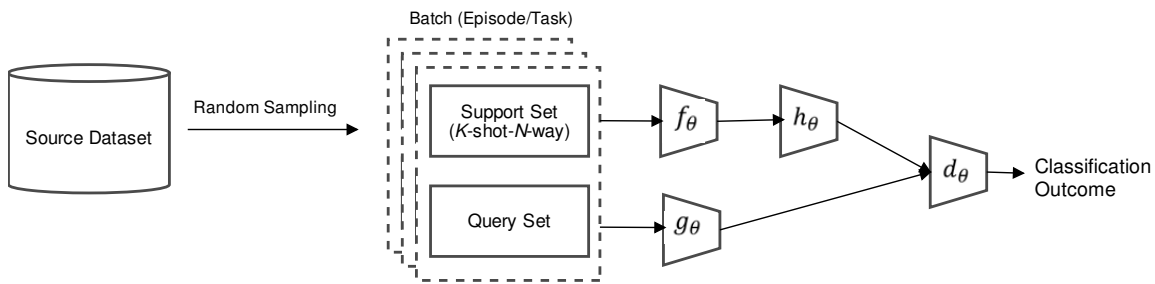}
  \caption{Metric-based FSL Architecture.}
  \label{fig:metric-based_FSL_arch}
\end{center}  
\end{figure}

%\begin{figure}[htp]
%\begin{center}
%    \includegraphics[width=0.3\textwidth]{Conv4_embedding_network.eps}
%    \caption{Conv-4 embedding network.}
%    \label{fig:conv-4_embedding_network}
%\end{center}
%\end{figure}

\subsection{Metric-based FSL}

This work summarized previous works that used metric-based approaches to deal with FSL problems and Fig.~\ref{fig:metric-based_FSL_arch} shows the overall architecture for these methods. A metric-based FSL model typically comprises three key components, including embedding functions ($f_\theta$ is for support set and $g_\theta$ is for query set), class representative function ($h_\theta$), and the distance metric function ($d_\theta$) that is used to determine the similarity of support and query embeddings. We aim to use non-parametric model to determine the class assignment as the work conducted by \citet{chen19closerfewshot} has empirically shown that a simple distance-based classifier could achieve competitive performance with the state-of-the-art meta-learning methods. Thus, Fig.~\ref{fig:metric-based_FSL_arch} uses a distance metric function ($d_\theta$) to perform classification without considering parametric functions.

It is apparent that embedding functions are crucial to the subsequent classification results. A common approach is to use a simple network such as Conv-4 as the embedding function, but a deep neural network such as ResNet-32 or a fully conditional embedding network such as bi-LSTM could be the alternative. Moreover, $f_\theta$ and $g_\theta$ can be different as used in matching network~\citep{vinyals2016matching}, and be identical as used in prototypical networks~\citep{snell2017prototypical} and relation networks~\citep{sung2018learning}. 

In a $K$-shot-$N$-way task, each class comprises several embedding vectors when $K > 1$, and $h_\theta$ accounts for the generation of the class representative for each class. Given the embedding vectors for each class, the $h_\theta$ in prototypical network is an average operator and outputs the class mean of these embedding vectors. As for relation network, the $h_\theta$ is a sum operator. On the other hand, matching network focuses on 1-shot learning, namely, $K=1$. In this case, the $h_\theta$ is an identity operator that directly outputs the input embedding vector.     

The final step for the metric-based FSL is to perform classification, and $d_\theta$ can be a classifier or distance metric. Parametric models needs to be optimized to treat new examples. In contrast, non-parametric models such as nearest neighbor method can perform classification only depending on the chosen metric, explaining why many metric-based methods used nearest neighbor method with appropriate distance metric to perform classification. Among these methods, matching networks and prototypical networks used pre-defined metrics, in which the former used cosine similarity, whereas the latter used Euclidean distance. Relation networks used a different approach by learning a distance metric from the relation module. 

%\begin{figure*}
%\centering
%\begin{minipage}{.5\textwidth}
%  \includegraphics[width=0.8\textwidth]{classifier.eps}
%  \caption{The architecture of classifier module.}
%\end{minipage}%
%\centering
%\begin{minipage}{.5\textwidth}
%  \includegraphics[width=0.8\textwidth]{Conv4_embedding_network.eps}
%  \caption{Conv-4 embedding network.}
%\end{minipage}
%\end{figure*}

%\begin{figure}
%  \includegraphics[width=0.3\textwidth]{Conv4_embedding_network.eps}
%  \caption{Conv-4 embedding network.}
%\end{figure}

%\begin{figure*}
%  \includegraphics{theta.eps}
%  \caption{Theta .}
%\end{figure*}

% Besides the three components mentioned above, constrastive loss and triplet loss
%\cite{yan2019dual} brings attention mechanism to the classifier.  In sum, most metric learning based approaches focus on data embedding network, the class proxies and the distance proxy.

%as shown in Fig.~\ref{fig:conv-4_embedding_network}

\subsection{Proxy Networks}
Based on the overall architecture introduced above, this work proposes a metric-based FSL method called ProxyNet. We propose to use the same embedding function to map support and query sets to an embedding space, namely, $f_\theta = g_\theta$. We follow the experimental protocol used in the work conducted by \citet{chen19closerfewshot} to use a Conv-4 network to conduct experiments and compared ProxyNet with other alternatives. Notably, different embedding functions can be used in our proposed model and we conduct experiments to evaluate the impact of different embedding networks on the performances. 
\begin{figure}
    \centering
    \includegraphics[width=1.0\textwidth]{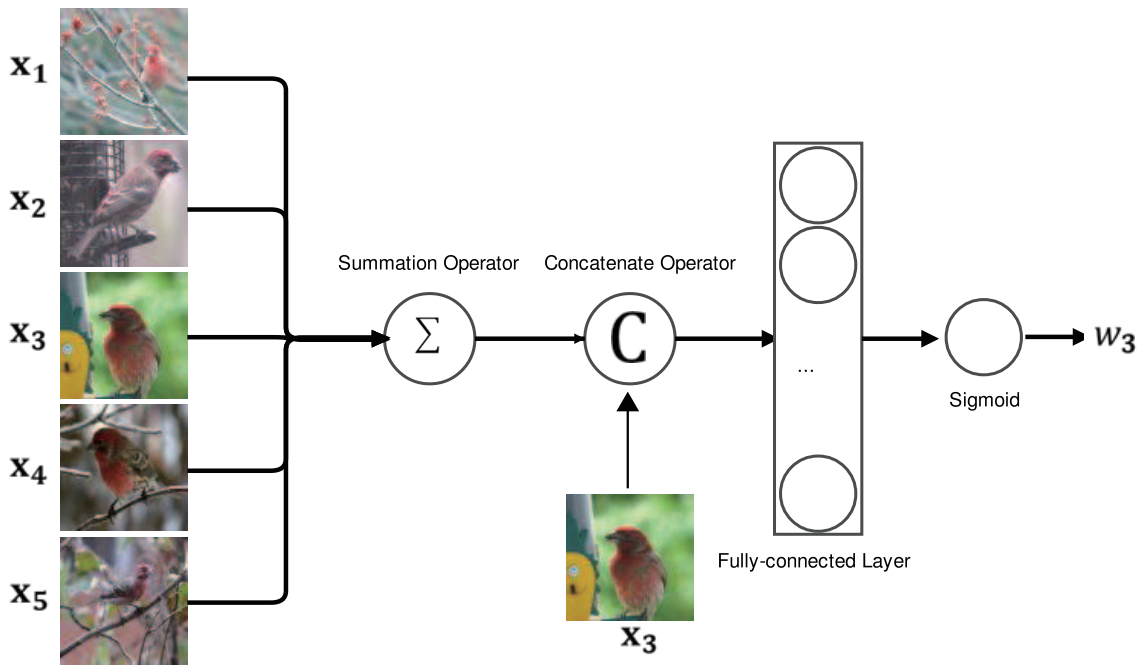}
    \caption{The architecture for the estimation of data weight in class proxy.}
    \label{fig:weight}
\end{figure}
Apart from the embedding functions, we propose to directly learn class representatives from data instead of using pre-defined operators such as average or sum to obtain the representatives. Given a class with $K$ points, $\mathbf{x}_1, \ldots, \mathbf{x}_K$, the center of the $K$ points is a commonly used point to represent the entire class. However, this approach will tend to be biased by outliers, so we propose a learning-based approach to learn the representative for each class.     
To simplify the notations for the following steps of our algorithm, the notations for the input images and their corresponding embeddings use the same notation $\mathbf{x}$. Given a $K$-shot-$N$-way task, there exists $\mathbf{x}_1^n, \ldots, \mathbf{x}_K^n$ examples for $n$th class, and we propose to use a function $h_\theta$ to learn the class representative of $n$th class $\mu_n$, namely, $\mu_n = h_\theta(\mathbf{x}_1^n, \ldots, \mathbf{x}_K^n)$, where $n \in \{1, \ldots, N)$. Following the same setting, one can obtain all the class representatives, $\mu_1, \ldots, \mu_N$, from $h_\theta$. We define the function as a weighted sum of the input points, namely, $h_\theta(\mathbf{x}_1^n, \ldots, \mathbf{x}_K^n) = w^n_1\mathbf{x}^n_1+w^n_2\mathbf{x}^n_2+ \ldots +w^n_K\mathbf{x}^n_K$, where $w^n_k$ ($k\in\{1,2, \ldots, K\}$) is the weight of the $k$th support point in the $n$th class. 
We propose to use a deep neural network as shown in Fig.~\ref{fig:weight} to learn the weight $w^n_k$, in which the inputs of the network comprises $\mathbf{x}^n_k$ and the summation of the data points for the $n$th class, say $\mathbf{s}^n$ . The goal is to learn the similarity score between $\mathbf{x}^n_k$ and $\mathbf{s}^n$, since $\mathbf{s}^n$ can be considered an approximation to the global trend of the data points in the $n$th class. Intuitively, the data point that is more close to the global trend of the class, more weight should be given to it. One advantage of such a design is to alleviate the influence of outliers by giving small weights to the outliers; in contrast, many pre-defined operators, such as sum or mean, are easily influenced by the outliers.
\begin{figure}
    \centering
    \includegraphics[width=1.0\textwidth]{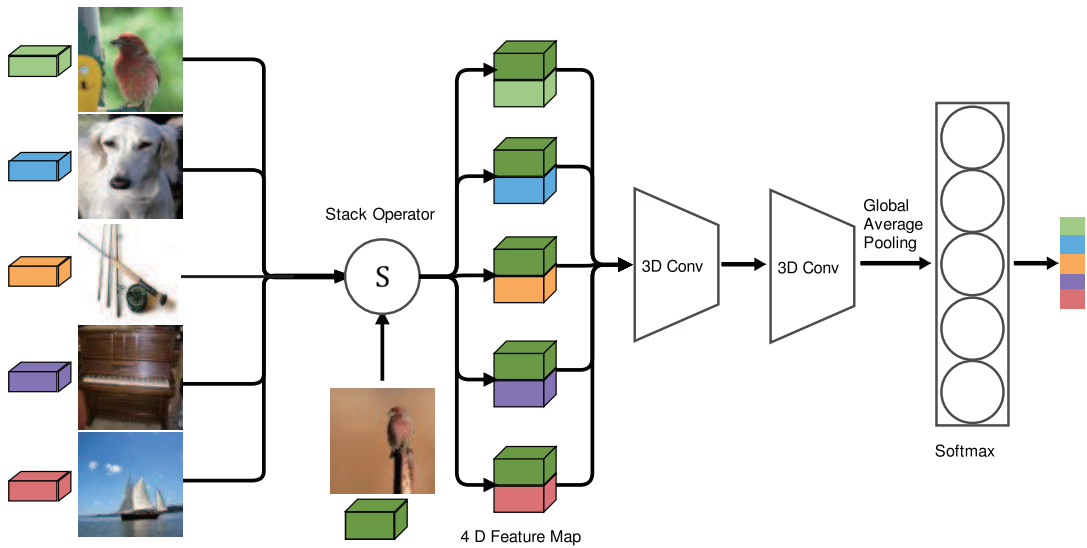}
    \caption{Classification module based on 3D Convolution.}
    \label{fig:classifier}
\end{figure}

Finally, metric-based approaches require to calculate the similarity between query data and class proxies, so that the class assignment can be obtained. Similarly, we propose to use a distance network~\citep{sung2018learning} to learn the similarity or distance between query data and each class proxy. Fig.~\ref{fig:classifier} shows the architecture of our classification module. Given the class proxies and a query embedding, we stack the query embedding and each class proxy embedding on a new dimension, so that the dimension is four after the stacking operator. Subsequently, we use a 3D convolutional network and global average pooling to obtain the probabilities of the class assignment. 

%First, the query embedding is concatenated with all class proxies one by one. Notably, we do not flatten the embeddings and the purpose is to retain the spatial feature of the embeddings. In this case, the embeddings comprise three dimensions, namely, channel, width, and height. Second, 

%Third, we concatenate the two embeddings along with the fourth dimension. Once the concatenation is completed, we use the distance network as shown in Figure~\ref{fig:classifier} to estimate the probabilities of class assignment. 

%Given a support embedding and a query embedding, the purpose of the distance network is to estimate the similarity of the two embeddings. 

As shown in Fig.~\ref{fig:classifier}, the 3D convolution network comprises two 3D convolutional blocks, each of which comprises 3D convolution, 3D batch normalization, and ReLU activation function. The aim is to learn the relation between channels and the features in each channel. Following the relation learning, a global average pooling layer is used to compute the average value for each feature map and supplies it to a softmax layer. Notably, we use a global average pooling layer to replace fully connected layers as used in \citep{lin2013network}, since it can decrease model complexity and reduce overfitting. Moreover, we list the number of trainable parameters for the proposed ProxyNet in the experimental section, showing the superiority of ProxyNet over state-of-the-art methods while requiring fewer trainable parameters. The output of the softmax layer comprises the probabilities of class assignments, and then the cross entropy loss function is used to train our proposed model. It is worth mentioning that the proposed model can be trained in an end-to-end manner. 
%$loss(\mathbf{x}, class) = -\log(\frac{\exp(x[class])}{\sum_{j}\exp(x[j])}) = -x[class] + \log(\sum_j \exp(x[j]))$

\section{Experiments}
This section introduces the datasets that used in the experiments, the experimental settings, and the experimental results on FSL problem.

\subsection{Datasets}
We evaluate the proposed model on two datasets, CUB~\citep{WahCUB_200_2011} and mini-ImageNet~\citep{vinyals2016matching}. Notably, these two datasets are commonly used by other previous FSL works. The CUB dataset comprises 100 classes for training, 50 classes for validation, and 50 classes for testing. This setting is the same as the evaluation protocol of \citep{hilliard2018few}. The mini-ImagNet dataset consists of 64 classes for training, 16 classes for validation, and 20 classes for testing. Each class has 600 images, each of which is of the size $84 \times 84$. Accuracy is used as the evaluation metric in the experiments.
% To keep the spatial features, we do not flat the feature map to an one dimension vector. Instead, we keep the three dimensions feature map, so as to feed the encoded features to the classifier module. 

\subsection{Experimental Settings}
We follow the experiment settings listed in \citep{chen19closerfewshot} to conduct experiments on mini-ImagNet and CUB datasets. First, all the methods use Conv-4 as the embedding network, which is consisted of 4 convolution layers and each convolution layer has 64 filters of size $3 \times 3$, since the purpose is to focus on the metric-based models on FSL problem rather than the effectiveness of the embedding network. Second, we evaluate all the methods on two scenarios, 1-shot-5-way and 5-shot-5-way, as they are mostly used evaluation scenarios in FSL research. Third, we randomly generate 600 test tasks from testing set, and the number of query images in each class is 15. Thus, for a 5-way-5-shot classification problem, each task comprises 75 query images as each class contains 15 images. We use the validation set to determine the best models in the experiments, in which we randomly generate 600 tasks from the validation set. The optimizer in our proposed model is SGD with an initial learning rate of 0.1 and the learning rate is adjusted by a reduce-on-plateau~\citep{pytorchdocument} method during training. The training stage comprises 300 epochs for the CUB dataset and 600 epochs for the mini-ImageNet dataset, each of which has 100 episodes. 
We use the same prepossessing methods as \citet{chen19closerfewshot} did in their work. More specially, all the images in both datasets are firstly resized to $92 \times 92$ and random cropped to $84 \times 84$, and then further processed by color jitters and random horizontal flip. The mean accuracy of generated tasks and their 95\% confidence intervals are presented as the experimental results. We have released the source code of our model on github~\footnote{ProxyNet: https://github.com/buptxiaofeng/proxynet\_fsl}, and detailed hyper-parameters as well as settings are available in the source code. 
\subsection{Experimental Results}
We conduct experiments on two datasets, and compare the proposed ProxyNet with several alternatives. The proposed method is a metric-based method, so all comparison methods are metric-based methods, including MatchingNet, ProtoNet, RelationNet, category traversal module (CTM)~\citep{li2019finding}, deep nearest
neighbor neural network (DN4)~\citep{li2019revisiting}, and baseline++. The introduction for the first three comparison methods can refer to the previous section. Notably, these comparison methods use non-parametric approaches to determine the class assignment, and they can be modeled by the proposed framework as listed in Fig.~\ref{fig:metric-based_FSL_arch}.

CTM~\citep{li2019finding} leveraged a concentrator and a projector to learn a mask which can be applied to all the support and query samples to obtain discriminative features. 
%In the design of Concentrator, the feature maps are firstly fed into a convolution network to reduce feature map dimension, and then the samples' feature maps in each class are averaged to extract the commonality among instances within each class.
Instead of learning class representatives and comparing the query samples with class representatives, CTM compares all the support and query sample pairs and uses the average distance between a query sample and all the support samples within a class as the distance to that class. Notably, we implement the three components of CTM, including concentrator, projector, and reshaper, with three residual blocks for a fair comparison, and the purpose is to make the number of trainable parameters closer to other models. Besides, the implementation for CTM uses Conv-4 as the embedding network, and uses two convolution layers as the distance metric as used by RelationNet. 

DN4~\citep{li2019revisiting} treated the feature maps generated by the embedding network of each sample as a series of local descriptor and proposed a Image-to-Class module to calculate the image-to-class similarity. Given a query sample, the local descriptors are firstly used to find the $k$ nearest neighbors in a class. Subsequently, the sum of similarities between the local descriptors of the query sample and the $k$ nearest neighbors is used to measure the image-to-class similarity between the query sample and the class. Our implementation for DN4 is based on the settings used by other comparison methods. More specifically, we used Conv-4 as the embedding network and used random crop, color jitters and random horizontal flip as the augmentation methods during training. Notably, we used 1-nearest neighbors on the CUB dataset and 3-nearest neighbors on the mini-ImageNet dataset as these settings were used by the original DN4.

The baseline++ was proposed by \citet{chen19closerfewshot}, and was developed based on transfer learning of network pre-training and fine-tuning. In training stage, the tasks in the meta-training dataset were used to train the underlying embedding network and a classifier of parameter $\mathbf{W} =[\mathbf{w}_1, \mathbf{w}_2, \ldots, \mathbf{w}_C]$, where $C$ is the number of classes or ways in the task. The class assignment is based on the cosine similarity between the query set and $\mathbf{w}_c$ ($1 \leq c \leq C)$. 

The parameters $\mathbf{W} =[\mathbf{w}_1, \mathbf{w}_2, \ldots, \mathbf{w}_C]$ resembles the $C$ prototypes used by ProtoNet, but differs in that $\mathbf{W}$ is learned from data. In fine-tuning stage, they used the support sets in the meta-testing dataset to fine-tune the classifier. Among the three components listed in Fig.~\ref{fig:metric-based_FSL_arch}, baseline++ trains the underlying embedding network and the class representatives, but uses a pre-defined distance function, cosine. 

Table~\ref{table:FSL_experimental_results} shows the experimental results. Notably, the experimental results for baseline++, MatchingNet, ProtoNet, and RelationNet are obtained from the work conducted by \citet{chen19closerfewshot}, and we follow their experimental settings for a fair comparison, including data augmentation methods and the embedding networks. More specially, we resize the images to the size of $92 \times 92$ followed by cropping to the size of $84 \times 84$. Subsequently, all the images are further processed by color jitters and random horizontal flip. The experimental results listed in Table~\ref{table:FSL_experimental_results} indicate that the proposed method significantly outperforms baseline++, MatchingNet, ProtoNet, and RelationNet. Meanwhile, the proposed method also generally achieves better results than CTM and DN4, but the performance differences between CTM and ProxyNet are insignificant in the two problems of the CUB dataset. It is worth mentioning that the concentrator of CTM can be treated as part of the embedding network which makes the embedding network deeper than the others. The other components of CTM also employed deeper networks, resulting in more trainable parameters than the proposed method. We list the numbers of trainable parameters for all the models in Table ~\ref{table:number_of_trainable_parameters}, which indicates that the number of trainable parameters for CTM is much more than that of ProxyNet. We provide detailed analysis regarding the impact of deeper embedding networks on the performance of ProxyNet in the following section. We summarize the three components and the learning type of the metric-based models and the proposed method in Table~\ref{table:model_summary}, in which the transfer learning in the table means that Basline++ follows the typical paradigm of transfer learning to train the model.

\begin{table}
\caption{Experimental results on 1-shot-5-way and 1-shot-5-way tasks.}
\centering
\resizebox{\textwidth}{20mm}{
\begin{tabular}{ c c c c c }
 \toprule
\label{table:FSL_experimental_results}
  & \multicolumn{2}{c}{CUB} & \multicolumn{2}{c}{mini-ImageNet} \\ \midrule
  Method & 1-shot & 5-shot & 1-shot & 5-shot\\
  \midrule
 \texttt{ProxyNet} & $67.52\pm 0.97$ & $\boldsymbol{82.85\pm 0.60}$ & $\boldsymbol{52.95 \pm 0.76}$ & $\boldsymbol{71.02 \pm 0.62}$  \\ 
 \texttt{CTM~\citep{li2019finding}} & $\boldsymbol{68.61}\pm 0.95$& $82.72 \pm 0.58 $& $50.80 \pm 0.84$ & $67.79\pm 0.67$ \\ 
 \texttt{DN4~\citep{li2019revisiting}} & $52.01\pm 0.86$& $76.70 \pm 0.64$& $47.32 \pm 0.80$ & $69.22\pm 0.66$ \\
 \texttt{Baseline++~\citep{chen19closerfewshot}} & $60.53\pm 0.83$& $79.34 \pm 0.61 $& $48.24 \pm 0.75$ & $66.43\pm 0.63$ \\ 
 \texttt{MatchingNet~\citep{vinyals2016matching}} & $60.52\pm 0.88$& $75.29 \pm 0.75 $& $48.14 \pm 0.78$ & $63.48\pm 0.66$ \\ 
 \texttt{ProtoNet~\citep{snell2017prototypical}} & $50.46\pm 0.88$ & $76.39 \pm 0.64$ & $44.42 \pm 0.84$ & $64.24\pm0.72$ \\
 \texttt{RelationNet~\citep{sung2018learning}} & $62.34\pm 0.94$ & $77.84 \pm 0.68$ & $49.31 \pm 0.85$ &  $66.60\pm0.69$ \\
 \bottomrule
\end{tabular}}
\end{table}

\begin{table}
\caption{The number of trainable parameters.}
\centering
\begin{tabular}{ c c c c c c c}
 \toprule
\label{table:number_of_trainable_parameters}
  ProxyNet & CTM & DN4 & Baseline++ & MatchingNet & ProtoNet & RelationNet\\
  \midrule
 $165,171$ & $307,921$ & $112,832$ & $433,288$ & $138,176$ & $113,600$ & $229,201$  \\ 
 \bottomrule
\end{tabular}
\end{table}

\begin{table}[ht!]
\caption{ Summary of metric-based learning models and the proposed model.}
\centering
\begin{adjustbox}{width=1\textwidth}
\begin{tabular}{ c c c c c}
\toprule
\label{table:model_summary}
   Method & Embedding Network & Class Proxy & Distance Metric & Type\\
  \midrule
 \texttt{ProxyNet} & ConvNet & trainable proxy & 3D-Conv and global pooling & meta learning \\ 
 \texttt{DN4}& ConvNet & No Proxy & Image-to-Class module(Cosine Distance) & meta learning \\
 \texttt{Baseline++} & ConvNet & No Proxy & Cosine Distance &  transfer learning \\
 \texttt{MatchingNet} & LSTM & No Proxy & Cosine Distance & meta learning\\
 \texttt{ProtoNet} & ConvNet & Mean of support samples & Euclidean Distance & meta learning \\
 \texttt{RelationNet} & ConvNet & Sum of support samples & 2D-Conv and fully connected layer & meta learning \\
 \bottomrule
\end{tabular}
\end{adjustbox}
\end{table}

\par

\section{Analysis and Discussion}
The experimental results in the previous section show that the proposed method can benefit from the learnable proxies for class representatives and learnable distance function. Moreover, the previous experiments only use simple CNNs as the embedding networks, deeper CNNs can be also the alternatives. Thus, we provide detailed analysis and discussion regarding the underlying embedding networks, proxy for class representative, and proxy for distance function in this section.

\begin{table}[ht!]
\caption{Experimental results on 5-way tasks using different embedding networks with augmentation methods.}
\centering
\begin{adjustbox}{width=1\textwidth}
\begin{tabular}{ c c c c c}
\toprule
\label{table:discussion_embedding_networks}
 & \multicolumn{2}{c}{CUB} & \multicolumn{2}{c}{mini-ImageNet} \\
  \midrule
  Embedding Network & 1-shot & 5-shot & 1-shot & 5-shot\\
   \midrule
 \texttt{Conv-4} & $67.52\pm 0.97$ & $ 82.85\pm 0.60 $& $52.95 \pm 0.76$ & $70.35 \pm 0.63$ \\ 
 \texttt{Conv-6} & $68.16\pm 0.93$ & $ 83.57 \pm 0.58 $& $52.18 \pm 0.82$ & $69.91 \pm 0.62$ \\
 \texttt{ResNet-10} & $76.79 \pm 0.84$ & $88.02 \pm 0.52$& $58.16 \pm 0.87$ & $75.27\pm 0.65$ \\
 \texttt{ResNet-18} & $76.72\pm 0.90$ & $88.63 \pm 0.49$& $57.88 \pm 0.87$ &  $75.23\pm0.66$ \\
 \texttt{ResNet-34} & $77.70\pm 0.86$ & $87.05 \pm 0.52$&$58.54 \pm 0.85$ &  $75.90\pm0.61$ \\
 \bottomrule
\end{tabular}
\end{adjustbox}
\end{table}

\subsection{Embedding Network}

In the previous experiments, the embedding network is a simple embedding network, Conv-4. We investigate whether the ProxyNet could benefit from a deeper CNN, so we conduct experiments by applying ProxyNet with different embedding networks to the two datasets that have undergone the data augmentation process as mentioned in the previous section. Table~\ref{table:discussion_embedding_networks} shows the experimental results, indicating that ProxyNet can normally benefits from deeper networks to obtain better performances. Although this work focuses on the design of networks for class representatives and distance function, it is apparent that the design of embedding network is also crucial to the model performance. Notably, data augmentation is a commonly used technique that can enable practitioners to significantly increase the size and diversity of the data, making it possible to train a deeper network under the FSL settings.  

Besides, we conduct experiments to investigate whether ProxyNet can benefit from a deeper CNN when no data augmentation is applied to the two datasets. The purpose is to analyze the relationship between the choice of embedding networks and the available data sizes. The experimental results are listed in Table~\ref{table:no_augmentation_results}.

\begin{table}[ht!]
\caption{Experimental results on 5-way tasks using different embedding networks without augmentation methods.}
\centering
\begin{adjustbox}{width=1\textwidth}
\begin{tabular}{ c c c c c}
\toprule
\label{table:no_augmentation_results}
 & \multicolumn{2}{c}{CUB (No data augmentation) } & \multicolumn{2}{c}{mini-ImageNet (No data augmentation)} \\
  \midrule
  Embedding Network & 1-shot & 5-shot & 1-shot & 5-shot\\
   \midrule
 \texttt{Conv-4} & $63.95\pm 0.95$ & $ 79.41\pm 0.64 $& $52.10 \pm 0.89$ & $71.02 \pm 0.62$ \\ 
 \texttt{Conv-6} & $63.22\pm 0.91$ & $ 77.24 \pm 0.70   $& $52.29 \pm 0.82$ & $69.52 \pm 0.66$ \\
 \texttt{ResNet-10} &  $62.49 \pm 0.95$ & $76.15 \pm 0.75$& $53.27 \pm 0.82$ & $69.10\pm 0.67$ \\
 \texttt{ResNet-18} & $63.47\pm 0.95$ & $72.50 \pm 0.78$& $53.64 \pm 0.82$ &  $68.86\pm0.66$ \\
 \texttt{ResNet-34} & $62.68\pm 1.01$ & $71.83 \pm 0.81$&$52.69 \pm 0.81$ &  $68.47\pm0.66$ \\
 \bottomrule
\end{tabular}
\end{adjustbox}
\end{table}

The experimental results in Table~\ref{table:no_augmentation_results} show an opposite trend of performances against network depth, meaning that the ProxyNet with simple CNN can normally yield better performance than that with deep CNNs. We conjecture that ProxyNet can not always benefit from a deep embedding network owing to the overfitting problem. The ProxyNet is a flexible design as it learns embedding network, class proxy, and distance metric proxy from data. This brings the model more capacity to cope with FSL problem, but also increases the model parameters. Thus, ProxyNet is prone to overfitting problem when the available samples are insufficient. 

%reduces the potential to improve the performance by using embedding network with more depth and width because of overfitting problem caused by the limitation of data and other factors.

%First, learning with a deeper neural network does not always lead to a better performance as many factors may influence model performance, such as the amount of training data, network architectures, and even optimization methods.  
Moreover, the results in Table~\ref{table:no_augmentation_results} show that it seems like that overfitting problem is more serious in CUB dataset than mini-ImageNet. The CUB dataset is designed for fine-grained image classification problem, meaning that the domain difference between base classes and novel classes is relative smaller than that of mini-ImageNet. Thus, FSL problem on CUB dataset is easier than that on mini-ImageNet. Besides, it is apparent that 5-shot-5-way problem is easier than 1-shot-5 way problem. Thus, it is expected that 5-shot-5-way problem on CUB dataset is the easiest problem, whereas 1-shot-5-way problem on mini-ImageNet is the most difficult problem among the problems listed in Table~\ref{table:no_augmentation_results}. Based on the problem difficulties and the experimental results presented in Table~\ref{table:no_augmentation_results}, it seems like that the ProxyNet with a deep embedding network is prone to overfitting when coping with easy problems. 
%the easier ProxyNet become overfitting with the increasing the amount of parameters to the embedding network, especially when we do not use any augmentation method to prevent overfitting for a fair comparison. The performance declination in the setting of 5-shot-5way on CUB dataset is the most significant because that is the easiest  problem setting among the four settings and suffers overfitting most. For the  hardest 1-shot-5-way problem on mini-ImageNet dataset, the performance is improved slightly by increasing more depth and width to the backbone, but ResNet-34 is worse than ResNet-18, which means that in this problem setting ProxyNet suffers overfitting least. 

We conduct experiments to empirically demonstrate that ProxyNet can benefit from data augmentation to dramatically improve performance as compared with that does not use data augmentation. Moreover, the findings show that ProxyNet should provide sufficient training examples when using a deep CNN as the embedding network; otherwise, the model may suffer from overfitting problem. In the setting of FSL, data augmentation is an effective approach to increase the sample size and avoid the overfitting problem.  

\subsection{Class Proxy}
When the underlying distance metric is Euclidean distance, a typical approach to obtain a class representative is to apply the average operator to the class members, meaning that the center point can be the representative as used by ProtoNet. The sum operator is similar to average operator by summing the class members as used by RelationNet. Different from these two operators, we propose to use weighted average of the class members to obtain the class representative, but the weights are learned from a neural network. 

%We use the summation of the class members as the global trend and use the cosine similarity between each support point and the global trend to calculate the weight for each support point. This is similar to our proposed approach, but directly uses cosine similarity to estimate the weights. These three operators should be specified in advance of the model training, whereas we propose to use weighted average approach to obtain the representative and the weights are learned from a network.  

\begin{figure}[h]
\includegraphics[width=\linewidth]{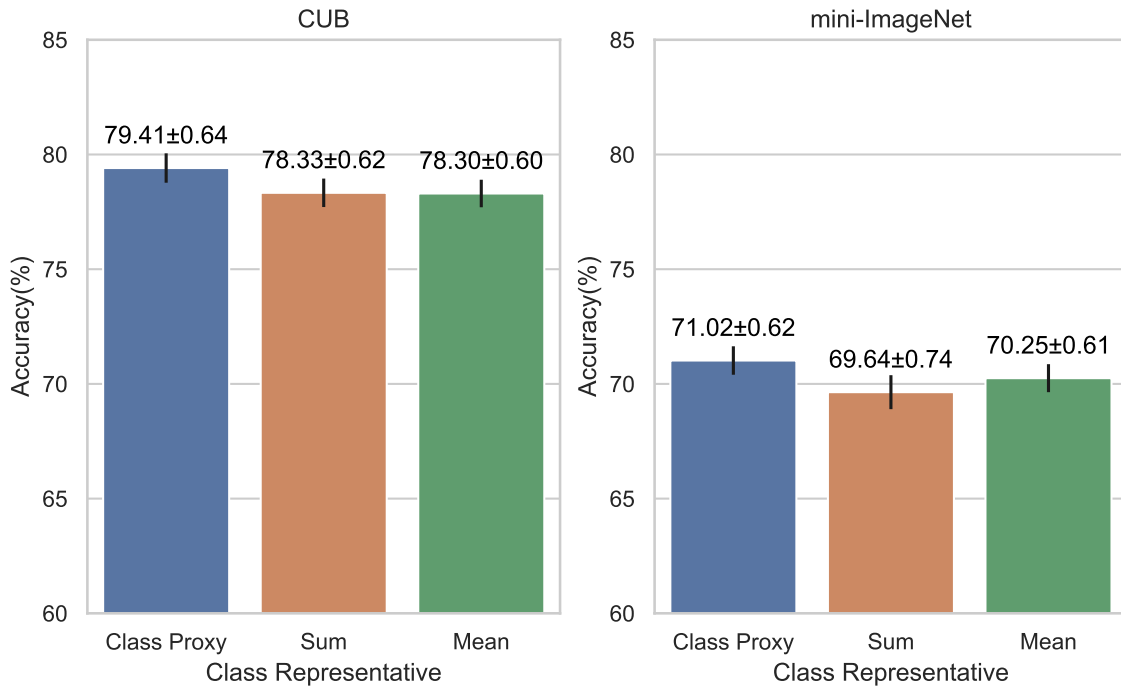}
\centering
\caption{Performance comparison with different approaches to obtain class representatives on 5-way-5-shot task.}
\label{fig:class_representative}
\end{figure}

To investigate the effectiveness of the proposed class representative proxy, we use our proposed method with mean and sum operators to obtain class representative as two comparison methods while fixing other settings. Fig.~\ref{fig:class_representative} shows the experimental results, indicating that the proposed class proxy yields better performance than the other two 
comparison methods. \citet{aggarwal2001surprising} has viewed the dimensionality curse from the point of view of the distance metrics and examined the behavior of the commonly used $L_k$ norm. They have shown that the problem of meaningfulness in high dimensionality is sensitive to the value of $k$ and the problem worsens faster for higher values of $k$. Thus, using Euclidean distance ($L_2$ norm) to search for the nearest neighbor is not always meaningful for high dimensional data. In contrast, using a network to directly learn class proxy is more preferable than using a pre-defined operator to obtain class representative.

\begin{figure}[h]
\includegraphics[width=\linewidth]{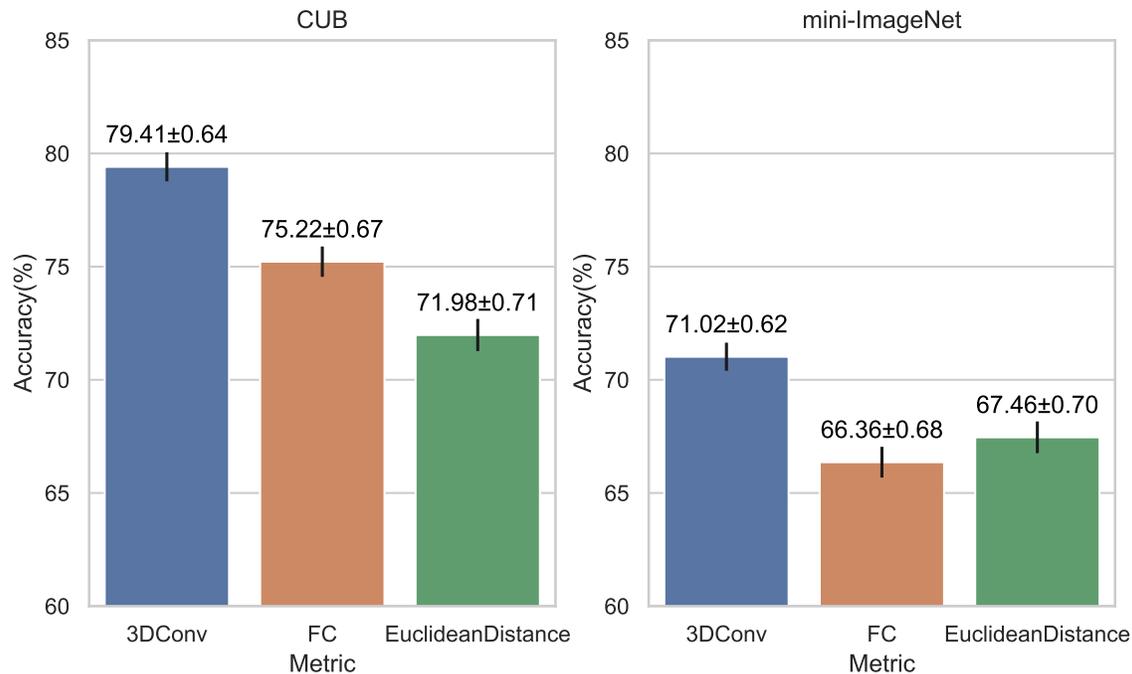}
\centering
\caption{Performance comparison using different metrics on 5-way-5-shot task.}
\label{fig:comparision_of_metrics}
\end{figure}

\subsection{Distance Metric Proxy}
The final step of metric-based methods is to determine the class assignment of the query embedding. Given support and query embeddings, the most commonly used approach is to measure the distances between query embedding and all the support embeddings. For example, ProtoNet uses Euclidean distance, while baseline++ uses cosine similarity. Instead of using fixed distance functions, RelationNet uses a relation module to learn the relation score from data, in which the network composed of a convolution layer, a max-pooling layer and a fully connected layer (FC in Fig.~\ref{fig:comparision_of_metrics}). This work also proposes to learn a distance function from data, but we propose to use a 3D convolutional network (3DConv in Fig.~\ref{fig:comparision_of_metrics}) to learn the relationship from data and use this kind of relationship as the proxy of distance metric.  

Fig.~\ref{fig:comparision_of_metrics} shows the experimental results on CUB and mini-ImageNet datasets, which indicate that the proposed ProxyNet yields the best performance among these methods. The ProxyNet uses a squzee-and-excition~\citep{hu2018squeeze} block, a 3D convolution, and a global average pooling layer to learn the relation scores between query embedding and support embeddings while retaining the spatial property. The model can benefit such a design by keeping the spatial feature and the relation of channels. 

This section provides detailed analysis and discussion about ProxyNet, and conduct extensive experiments to empirically show that each component of our proposed model is crucial to the model architecture. Moreover, using a data-driven approach to learn the embeddings, class proxy, and distance metric proxy can directly learn the three components from data without human intervention, giving a base to develop a flexible and generalized model.  

%Distance metric is the last component of metric based model. Popular choices of distance metric includes euclidean distance, cosine, both of which are fixed distance functions. On the other hand, the proposed distance metric and metric in the RelationNet are both trainable. The main difference between these two trainable distance metric is the design of architecture. In RelationNet, the distance metric (FC Distance) is composed of convolution layer, max pooling layer and fully connected layer.

\section{Conclusions} 
This work analyzes the previous metric-learning based model and summarizes three important components in a metric-learning based model. Based on our analysis, we propose a metric-learning based model called ProxyNet to cope with FSL problem. Central to the proposed model is to use data-driven approach to consider embeddings, class proxy, and distance metric proxy as three learning problems, making it possible to learn these components from data without human intervention. The experimental results indicate that the proposed method can yield promising results on CUB and mini-ImageNet datasets in 1-shot-5-way and 5-shot-5-way scenarios. Detailed analysis and discussion about ProxyNet are also provided in this work. The future work is to cope with FSL problem that comprises cross-domain datasets. 

\section*{Acknowledgment}
This work was supported in part by Ministry of Science and Technology, Taiwan, under Grant no. MOST 107-2221-E-009-109-MY2 and MOST 109-2628-E-009-009-MY3. We are grateful to the National Center for High-performance Computing for computer time and facilities.

%\acks{Acknowledgements should go at the end, before appendices and references.}

%\bibliographystyle{plain}
%\bibliography{acml19}
\bibliography{reference}

%\appendix

%\section{First Appendix}\label{apd:first}

%This is the first appendix.

%\section{Second Appendix}\label{apd:second}

%This is the second appendix.

\end{document}